\documentclass[10pt,twocolumn,letterpaper]{article}

\usepackage{iccv}
\usepackage{times}
\usepackage{epsfig}
\usepackage{graphicx}
\usepackage{amsmath}
\usepackage{enumitem}
\usepackage{amssymb}
\usepackage{xcolor}
\usepackage{booktabs}
\usepackage{algorithmic}
\usepackage{verbatim}

\newcommand\swish{\operatorname{swish}}
\newcommand{\sigmoid}{\operatorname{\sigma}}
\newcommand{\hardswish}{\operatorname{h-swish}}

\usepackage[pagebackref=true,breaklinks=true,colorlinks,bookmarks=false]{hyperref}

\iccvfinalcopy %

\def\iccvPaperID{5162} %

\newcommand{\omt}[1]{}

\ificcvfinal\fi
\begin{document}

\title{Searching for MobileNetV3}

\author{
	\small
	\begin{tabular}{c c c c c c c}                              
	    \bf Andrew Howard$^1$ &
	    \bf Mark Sandler$^1$ &
	    \bf Grace Chu$^1$ &
	    \bf Liang-Chieh Chen$^1$ &
	    \bf Bo Chen$^1$ &
		\bf Mingxing Tan$^2$ \\
		\bf Weijun Wang$^1$ &
		\bf Yukun Zhu$^1$ &
		\bf Ruoming Pang$^2$ &
		\bf Vijay Vasudevan$^2$ &
		\bf Quoc V. Le$^2$ &
		\bf Hartwig Adam$^1$\\                                
		\multicolumn{7}{c}{$^1$Google AI, $^2$Google Brain} \\                                              
		\multicolumn{7}{c}{\{howarda, sandler, cxy, lcchen, bochen, tanmingxing, weijunw, yukun, rpang, vrv, qvl, hadam\}@google.com} \\
	\end{tabular}                                                                       
} 

\maketitle
\ificcvfinal\thispagestyle{empty}\fi
\begin{abstract}
 We present the next generation of MobileNets based on a combination of complementary search techniques as well as a novel architecture design. MobileNetV3 is tuned to mobile phone CPUs through a combination of hardware-aware network architecture search (NAS) complemented by the NetAdapt algorithm and then subsequently improved through novel architecture advances. This paper starts the exploration of how automated search algorithms and network design can work together to harness complementary approaches improving the overall state of the art.  Through this process we create two new MobileNet models for release: MobileNetV3-Large and MobileNetV3-Small which are targeted for  high and low resource use cases. These models are then adapted and applied to the tasks of object detection and semantic segmentation. For the task of semantic segmentation (or any dense pixel prediction), we propose a new efficient segmentation decoder Lite Reduced Atrous Spatial Pyramid Pooling (LR-ASPP). We achieve new state of the art results for mobile classification, detection and segmentation. MobileNetV3-Large is 3.2\% more accurate on ImageNet classification while reducing latency by 20\% compared to MobileNetV2. MobileNetV3-Small is 6.6\% more accurate compared to a MobileNetV2 model with comparable latency. MobileNetV3-Large detection is over 25\% faster at roughly the same accuracy as MobileNetV2 on COCO detection. MobileNetV3-Large LR-ASPP is 34\% faster than MobileNetV2 R-ASPP at similar accuracy for Cityscapes segmentation. 

\end{abstract}

\section{Introduction}

\begin{figure}[!t]
    \centering
    \includegraphics[width=0.47\textwidth]{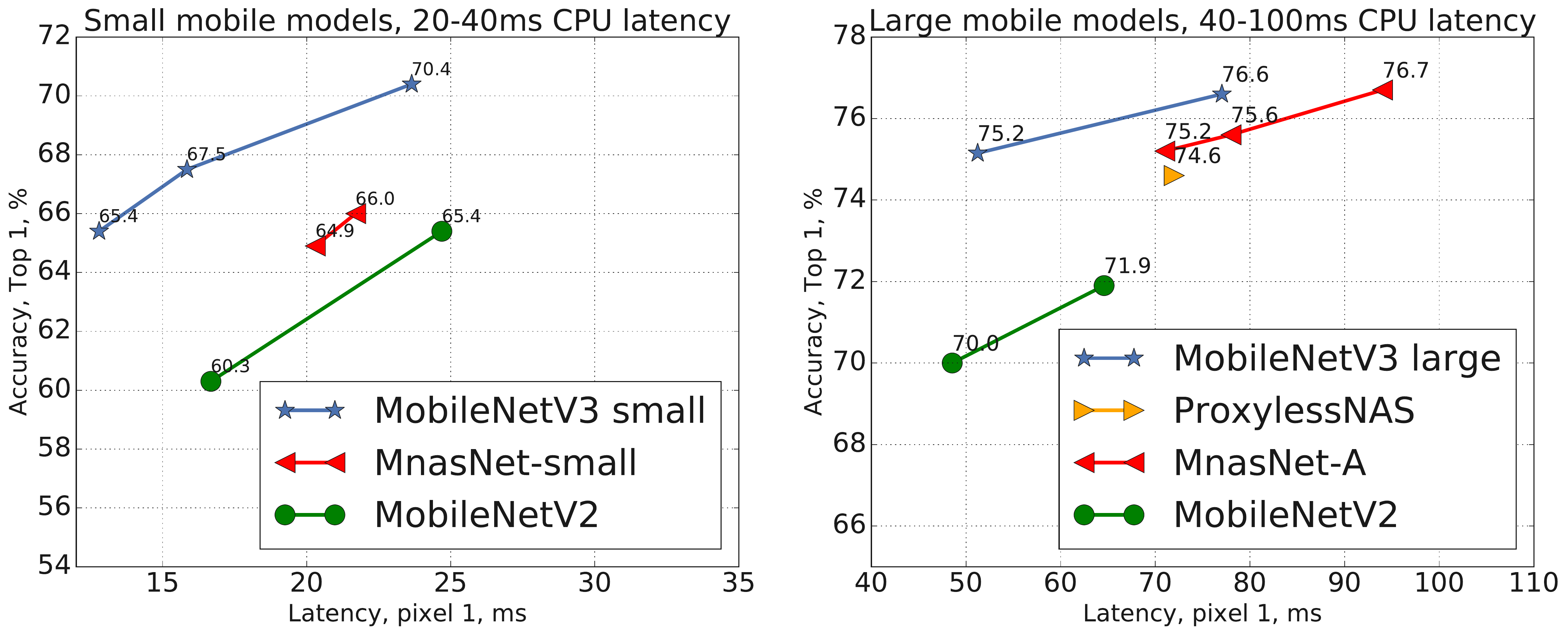}
    \caption{The trade-off between Pixel 1 latency and top-1 ImageNet accuracy. All models use the input resolution 224. V3 large and V3 small use multipliers 0.75, 1 and 1.25 to show optimal frontier. All latencies were measured on a single large core of the same device using TFLite\cite{tensorflow2015-whitepaper}. MobileNetV3-Small and Large are our proposed next-generation mobile models.}
    \label{fig:latency_vs_top_1}
\end{figure}
\begin{figure}[!t]
    \centering
    \includegraphics[width=0.47\textwidth]{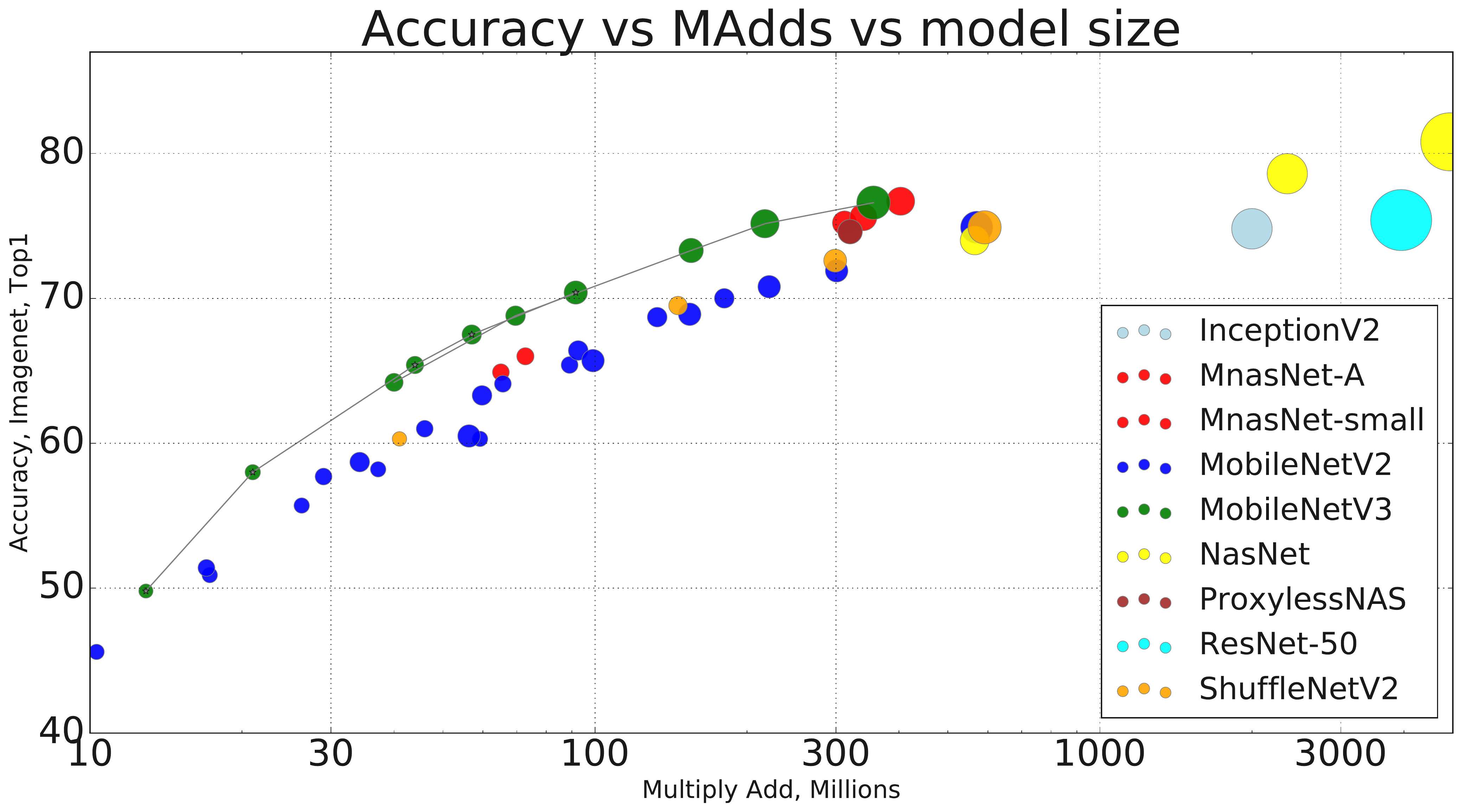}
    \caption{The trade-off between MAdds and top-1 accuracy. This allows to compare models that were targeted different hardware or software frameworks. All MobileNetV3 are for input resolution 224 and use multipliers 0.35, 0.5, 0.75, 1 and 1.25. See section \ref{section:experiments} for other resolutions. Best viewed in color.}
    \label{fig:madds_vs_top_1}
\end{figure}

Efficient neural networks are becoming ubiquitous in mobile applications enabling entirely new on-device experiences. They are also a key enabler of personal privacy allowing a user to gain the benefits of neural networks without needing to send their data to the server to be evaluated. Advances in neural network efficiency not only improve user experience via higher accuracy and lower latency, but also help preserve battery life through reduced power consumption.

This paper describes the approach we took to develop MobileNetV3 Large and Small models in order to deliver the next generation of high accuracy efficient neural network models to power on-device computer vision. The new networks push the state of the art forward and demonstrate how to blend automated search with novel architecture advances to build effective models.

The goal of this paper is to develop the best possible mobile computer vision architectures optimizing the accuracy-latency trade off on mobile devices. To accomplish this we introduce (1) complementary search techniques, (2) new efficient versions of nonlinearities practical for the mobile setting, (3) new efficient network design, (4) a new efficient segmentation decoder. We present thorough experiments demonstrating the efficacy and value of each technique evaluated on a wide range of use cases and mobile phones.

The paper is organized as follows. We start with a discussion of related work in Section \ref{section:related}. Section \ref{section:building} reviews the efficient building blocks used for mobile models. Section \ref{section:search} reviews architecture search and the complementary nature of MnasNet and NetAdapt algorithms. Section \ref{section:improvements} describes novel architecture design improving on the efficiency of the models found through the joint search. Section \ref{section:experiments} presents extensive experiments for classification, detection and segmentation in order do demonstrate efficacy and understand the contributions of different elements. Section \ref{section:conclusions} contains conclusions and future work.

\section{Related Work}
\label{section:related}
Designing deep neural network architecture for the optimal trade-off between accuracy and efficiency has been an active research area
in recent years. Both novel hand-crafted structures and algorithmic neural architecture search have played important roles in advancing this field.

SqueezeNet\cite{SqueezeNet} extensively uses $1$x$1$ convolutions with squeeze and expand modules primarily focusing on reducing the number of parameters. More recent works shifts the focus from reducing parameters to reducing the number of operations (MAdds) and the actual measured latency. MobileNetV1\cite{MobilenetV1} employs depthwise separable convolution to substantially improve computation efficiency. MobileNetV2\cite{mobilenetv2} expands on this by introducing a resource-efficient block with inverted residuals and linear bottlenecks. ShuffleNet\cite{ShuffleNet2017} utilizes group convolution and channel shuffle operations to further reduce the MAdds. CondenseNet\cite{CondenseNet} learns group convolutions at the training stage to keep useful dense connections between layers for feature re-use. ShiftNet\cite{Shift} proposes the shift operation interleaved with point-wise convolutions to replace expensive spatial convolutions.

To automate the architecture design process, reinforcement learning (RL) was first introduced to search efficient architectures with competitive accuracy \cite{NAS_reinforcement,LearningToLearnScale,BakerGNR16,ProgressiveNAS,ENAS}. A fully configurable search space can grow exponentially large and intractable. So early works of architecture search focus on the cell level structure search, and the same cell is reused in all layers. Recently, \cite{mnasnet} explored a block-level hierarchical search space allowing different layer structures at different resolution blocks of a network. To reduce the computational cost of search, differentiable architecture search framework is used in \cite{DARTS,proxyless,FBnet} with gradient-based optimization. Focusing on adapting existing networks to constrained mobile platforms, \cite{NetAdapt,AMC,ChamNet} proposed more efficient automated network simplification algorithms.

Quantization \cite{fakequant,quantwhite,QCNN,SouHub14,Increment_quant,DoReFa-Net,Xnor-Net} is another important complementary effort to improve the network efficiency through reduced precision arithmetic. Finally, knowledge distillation \cite{ModelCompression,Distill} offers an additional complementary method to generate small accurate "student" networks with the guidance of a large "teacher" network.

\section{Efficient Mobile Building Blocks}
\label{section:building}
Mobile models have been built on increasingly more efficient building blocks. MobileNetV1 \cite{MobilenetV1} introduced {\it depthwise separable convolutions} as an efficient replacement for traditional convolution layers. Depthwise separable convolutions effectively factorize traditional convolution by separating spatial filtering from the feature generation mechanism. Depthwise separable convolutions are defined by two separate layers: light weight depthwise convolution for spatial filtering and heavier 1x1 pointwise convolutions for feature generation.

\begin{figure}[!t]
    \centering
    \includegraphics[width=0.47\textwidth]{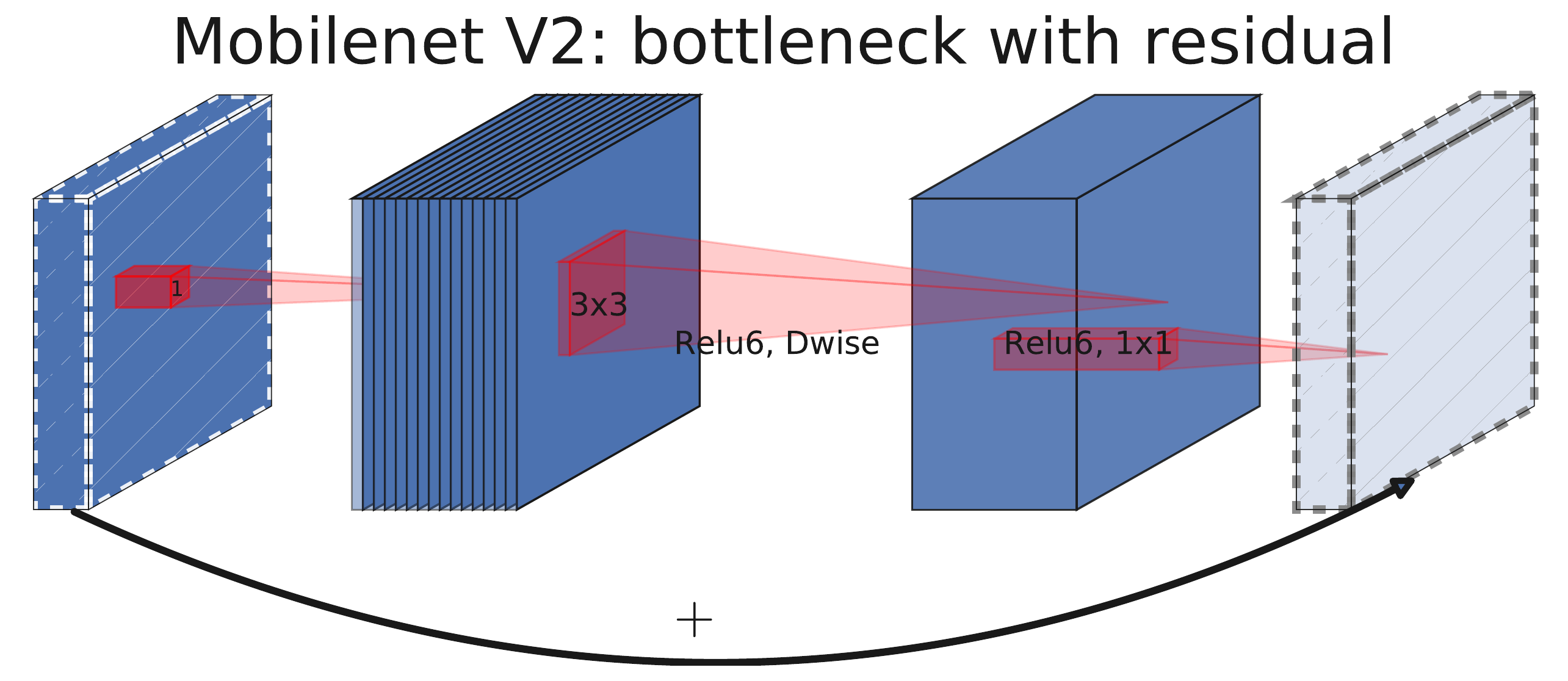}
    \caption{MobileNetV2 \cite{mobilenetv2} layer (Inverted Residual and Linear Bottleneck).
    Each block consists of narrow input and output (bottleneck),
    which don't have nonlinearity, followed by expansion to a much higher-dimensional space and
    projection to the output.
    The residual connects bottleneck (rather than expansion).}
    \label{fig:expansion_conv_v2}
\end{figure}

MobileNetV2~\cite{mobilenetv2} introduced the linear bottleneck and inverted residual structure in order to make even more efficient layer structures by leveraging the low rank nature of the problem. This structure is shown on Figure \ref{fig:expansion_conv_v2} and is defined by a 1x1 expansion convolution followed by depthwise convolutions and a 1x1 projection layer. The input and output are connected with a residual connection if and only if they have the same number of channels. This structure maintains a compact representation at the input and the output while expanding to a higher-dimensional feature space internally to increase the expressiveness of nonlinear per-channel transformations.

\begin{figure}[!t]
    \centering
    \includegraphics[width=0.47\textwidth]{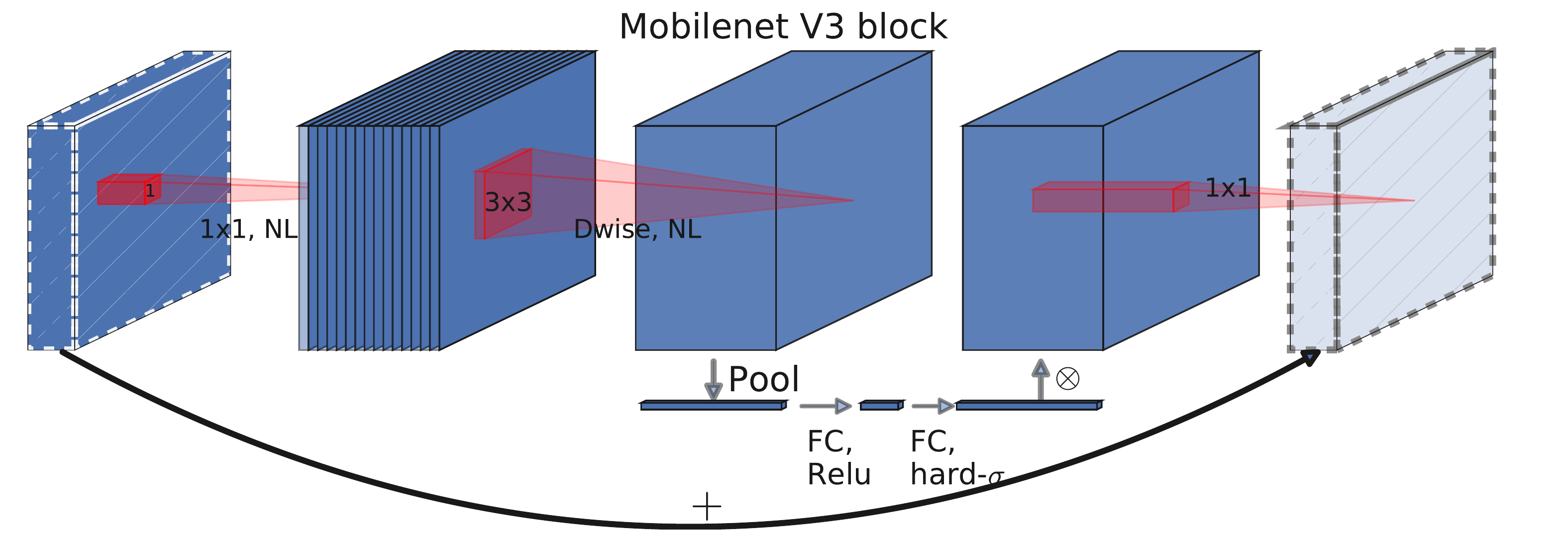}
    \caption{MobileNetV2 + Squeeze-and-Excite \cite{squeezeandexcite}. In contrast with \cite{squeezeandexcite} we apply the squeeze and excite in the residual layer.  We use different nonlinearity depending on the layer, see section \ref{section:nonlinearities} for details.}
    \label{fig:expansion_conv_v3}
\end{figure}

MnasNet~\cite{mnasnet} built upon the MobileNetV2 structure by introducing lightweight attention modules based on squeeze and excitation into the bottleneck structure. Note that the squeeze and excitation module are integrated in a different location than ResNet based modules proposed in \cite{squeezeandexcite}. The module is placed after the depthwise filters in the expansion in order for attention to be applied on the largest representation as shown on Figure \ref{fig:expansion_conv_v3}.

For MobileNetV3, we use a combination of these layers as building blocks in order to build the most effective models. Layers are also upgraded with modified $\swish$ nonlinearities \cite{swish, silu, gelu}. Both squeeze and excitation as well as the swish nonlinearity use the sigmoid which can be inefficient to compute as well challenging to maintain accuracy in fixed point arithmetic so we replace this with the hard sigmoid \cite{hard-swish,binaryconnect} as discussed in section \ref{section:nonlinearities}.

\section{Network Search}
\label{section:search}
Network search has shown itself to be a very powerful tool for discovering and optimizing network architectures \cite{NAS_reinforcement, mnasnet, proxyless, NetAdapt}. For MobileNetV3 we use platform-aware NAS to search for the global network structures by optimizing each network block. We then use the NetAdapt algorithm to search per layer for the number of filters. These techniques are complementary and can be combined to effectively find optimized models for a given hardware platform.   

\subsection{Platform-Aware NAS for Block-wise Search}

\omt{
We employ a similar factorized search space as \cite{mnasnet} to explore the initial model candidates. Specifically, we partition the network into a list of predefined stages, and search for layer architectures for each stage respectively. However, unlike MnasNet \cite{mnasnet} that restricts all layers to be the same in each stage, we also search the kernel size and expansion ratio for each individual layer in each stage, thus permitting more layer diversity throughout the
network. Similarly to \cite{mnasnet}, we use MobileNetV2~\cite{mobilenetv2} as a reference, and
search for convolutional op types, expansion ratios (3 or 6), kernel sizes (3x3 or 5x5),
squeeze-and-excite ratio (0.5, 0.25, or 0.125). For each stage, we also search for the number of
layers (0, +1, or -1) and the filter sizes with ratio of (0.75, 1.0, or 1.2) respective to MobileNetV2~
\cite{mobilenetv2}

Similar to MnasNet \cite{mnasnet}, we use a RNN-based controller
with the same multi-objective reward $ACC(m) \times \left[LAT(m)/T \right] ^ w$, which approximates the Pareto-optimal solutions by balancing model accuracy $ACC(m)$ and latency $LAT(m)$ for each model $m$ based on the target latency $T$. For big models with 75ms target latency, we use the same coefficients $w=-0.07$ as \cite{mnasnet}, but for small models, we observe that accuracy changes more dramatically with latency; therefore, we use a smaller co-efficient $w=-0.15$ to compensate for the larger accuracy change for different latencies.
}

Similar to~\cite{mnasnet}, we employ a platform-aware neural architecture approach to find the global network structures. Since we use the same RNN-based controller and the same factorized hierarchical search space, we find similar results as \cite{mnasnet} for Large mobile models with target latency around 80ms. Therefore, we simply reuse the same MnasNet-A1 \cite{mnasnet} as our initial Large mobile model, and then apply NetAdapt \cite{NetAdapt} and other optimizations on top of it.

However, we observe the original reward design is not optimized for small mobile models. Specifically, it uses a multi-objective reward $ACC(m) \times \left[LAT(m)/TAR \right] ^ w$ to approximate Pareto-optimal solutions, by balancing model accuracy $ACC(m)$ and latency $LAT(m)$ for each model $m$ based on the target latency $TAR$. We observe that accuracy changes much more dramatically with latency for small models; therefore, we need a smaller weight factor $w=-0.15$ (vs the original $w=-0.07$ in \cite{mnasnet}) to compensate for the larger accuracy change for different latencies. Enhanced with this new weight factor $w$, we start a new architecture search from scratch to find the initial seed model  and then apply NetAdapt and other optimizations to obtain the final MobileNetV3-Small model.

\subsection{NetAdapt for Layer-wise Search}
The second technique that we employ in our architecture search is NetAdapt~\cite{NetAdapt}. This 
approach is complimentary to platform-aware NAS: it allows fine-tuning of individual layers in a sequential manner,
rather than trying to infer coarse but global architecture. We refer to the original paper for the full details.
In short the technique proceeds as follows:
\begin{enumerate}[itemsep=0mm]
\item Starts with a seed network architecture found by platform-aware NAS. 
\item For each step: 
\begin{enumerate}[itemsep=0mm]
\item Generate a set of  new {\it proposals}. Each proposal
represents a modification of an architecture that generates at least $\delta$ reduction in latency compared to the previous step.
\item For each proposal we use the pre-trained model from the  previous step and populate the new proposed architecture, truncating and randomly initializing missing
weights as appropriate. Fine-tune each proposal for $T$ steps to get a coarse estimate of the accuracy.
\item  Selected best proposal according to some metric. 
\end{enumerate}
\item Iterate previous step until target latency is reached.
\end{enumerate}
 In \cite{NetAdapt} the metric was to minimize the accuracy change. 
We modify this algorithm and minimize the ratio between latency change and accuracy change. That is for all proposals generated during each NetAdapt step, we pick one that maximizes:  $\frac{\Delta\text{Acc}}{|\Delta\text{latency}|},$  with $\Delta\text{latency}$ satisfying the constraint in 2(a). The intuition is that because our proposals are discrete, we prefer proposals that maximize the slope of the trade-off curve.

This process is repeated until the latency reaches its target,  and then we re-train the new architecture from scratch.
We use the same proposal generator as was used in ~\cite{NetAdapt} for MobilenetV2. Specifically, we allow the following two  types of proposals: 
\begin{enumerate}
    \item Reduce the size of any expansion layer;
    \item Reduce bottleneck in all blocks that share the same bottleneck size - to maintain residual connections.
\end{enumerate}
For our experiments we used $T=10000$ and find that while it increases the accuracy of the initial fine-tuning of the proposals, it does not however, change the final accuracy when trained from scratch. We set $\delta=0.01 | L |$, where $L$ is the latency of the seed model. 

\section{Network Improvements}
\label{section:improvements}
In addition to network search, we also introduce several new components to the model to further improve the final model. We redesign the computionally-expensive layers at the beginning and the end of the network. We also introduce a new nonlinearity, h-swish, a modified version of the recent swish nonlinearity, which is faster to compute and more quantization-friendly.

\subsection{Redesigning Expensive Layers}

Once models are found through architecture search, we observe that some of the last layers as well as some of the earlier layers are more expensive than others. We propose some modifications to the architecture to reduce the latency of these slow layers while maintaining the accuracy. These modifications are outside of the scope of the current search space. 

The first modification reworks how the last few layers of the network interact in order to produce the final features more efficiently. Current models based on MobileNetV2's inverted bottleneck structure and variants use 1x1 convolution as a final layer in order to expand to a higher-dimensional feature space. This layer is critically important in order to have rich  features for prediction. However, this comes at a cost of extra latency. 

To reduce latency and preserve the high dimensional features, we move this layer past the final average pooling. This final set of features is now computed at 1x1 spatial resolution instead of 7x7 spatial resolution. The outcome of this design choice is that the computation of the features becomes nearly free in terms of computation and latency.

\begin{figure}[!t]
    \centering
    \includegraphics[width=0.47\textwidth]{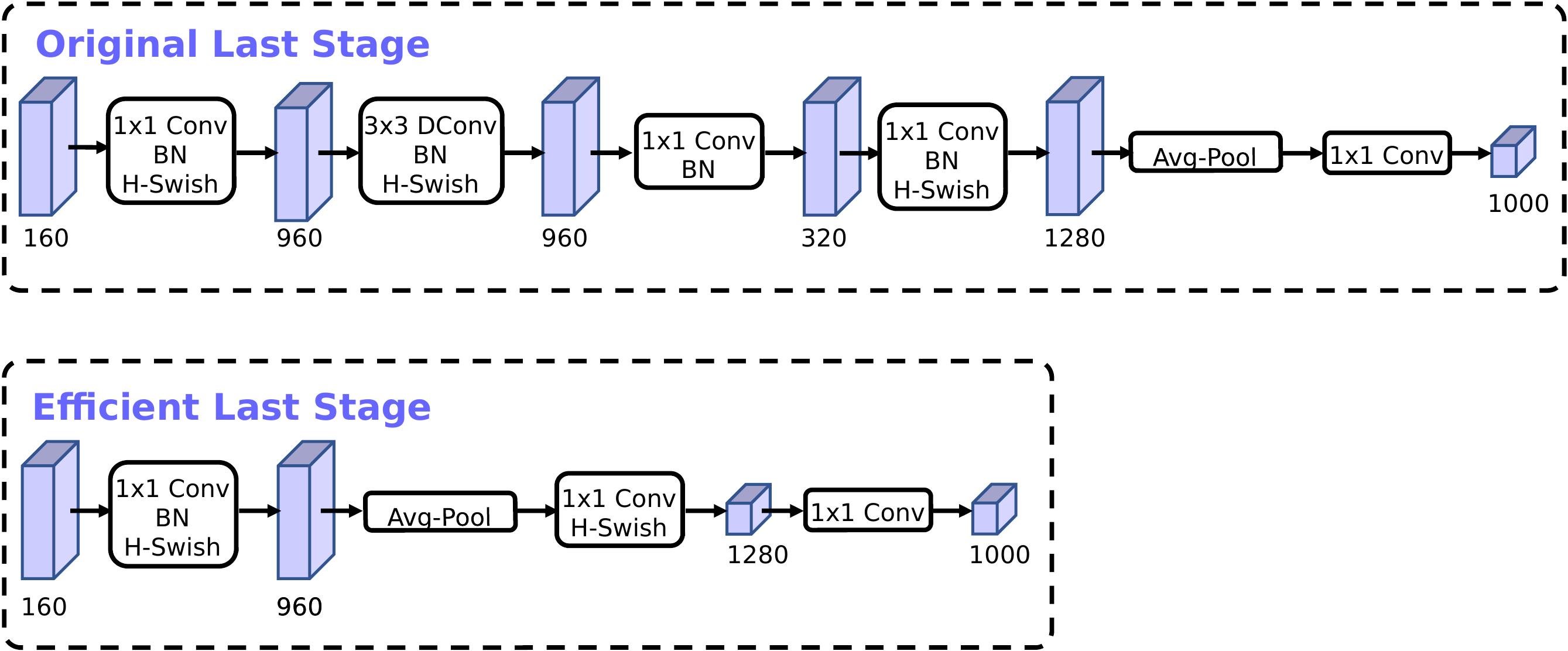}
    \caption{Comparison of original last stage and efficient last stage. This more  efficient last stage is able to drop three expensive layers at the end of the network at no loss of accuracy.}
    \label{fig:mobilenetv3_last_stage}
\end{figure}

Once the cost of this feature generation layer has been mitigated, the previous bottleneck projection layer is no longer needed to reduce computation. This observation allows us to remove the projection and filtering layers in the previous bottleneck layer, further reducing computational complexity. The original and optimized last stages can be seen in figure \ref{fig:mobilenetv3_last_stage}. The efficient last stage reduces the latency by 7 milliseconds which is 11\% of the running time and reduces the number of operations by 30 millions MAdds with almost no loss of accuracy. Section \ref{section:experiments} contains detailed results. 

Another expensive layer is the initial set of filters. Current mobile models tend to use 32 filters in a full 3x3 convolution to build initial filter banks for edge detection. Often these filters are mirror images of each other. We experimented with reducing the number of filters and using different nonlinearities to try and reduce redundancy. We settled on using the hard swish nonlinearity for this layer as it performed as well as other nonlinearities tested. We were able to reduce the number of filters to 16 while maintaining the same accuracy as 32 filters using either ReLU or swish. This saves an additional 2 milliseconds and 10 million MAdds.

\subsection{Nonlinearities}
\label{section:nonlinearities}
\newcommand\relus{\operatorname{ReLU6}}
\newcommand\relu{\operatorname{ReLU}}

In \cite{swish, silu, gelu} a nonlinearity called {\it swish} was introduced that when used as a drop-in replacement for $\relu$, that significantly improves the accuracy of neural networks.  The nonlinearity is defined as $$
 \swish x = x \cdot \sigmoid(x)
$$
While this nonlinearity improves accuracy, it comes with non-zero cost in embedded environments as the sigmoid function is much more expensive  to compute on mobile devices.
We deal with this problem in two ways. 

\begin{figure}[!t]
    \centering
    \includegraphics[width=0.47\textwidth]{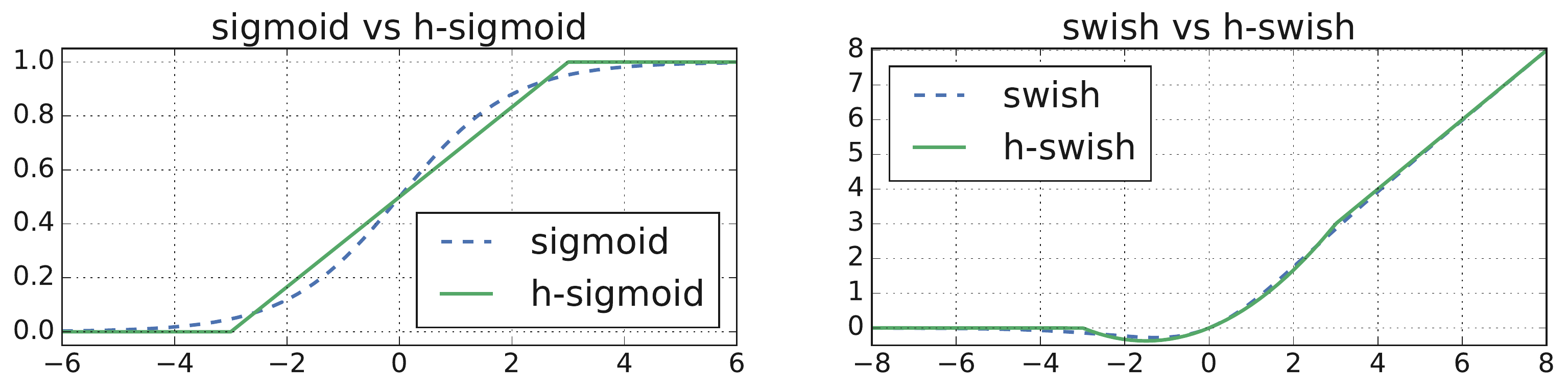}
    \caption{Sigmoid and swish nonlinearities and ther ``hard'' counterparts.}
    \label{fig:qsigm}
\end{figure}

\paragraph{1.} We replace sigmoid function with its piece-wise linear hard analog: 
$\frac{\relus(x + 3)}{6}$ similar to \cite{binaryconnect,spse1992curves}. The minor difference is we use $\relus$ rather than a custom clipping constant.  Similarly, the hard version of swish becomes $$
\hardswish [x] = x \frac{\relus(x + 3)}{6}
$$
A similar version of hard-swish was also recently proposed in~ \cite{hard-swish}. The comparison of the soft and hard version of sigmoid and swish nonlinearities is shown in figure \ref{fig:qsigm}. Our choice of constants was motivated by simplicity and
being a good match to the original smooth version. 
In our experiments, we found hard-version of all these functions to have no discernible difference in accuracy, but multiple advantages from a deployment perspective.  First, optimized implementations of $\relus$ are available  on virtually all software and hardware frameworks.
Second, in quantized mode, it eliminates potential numerical precision loss caused by different implementations of the approximate sigmoid.  Finally, in practice, h-swish can be implemented as a piece-wise function to reduce the number of memory accesses driving the latency cost down substantially. 
\paragraph{2.} The cost of applying nonlinearity decreases as we go deeper into the network, since
each layer activation memory typically halves every time the resolution drops. Incidentally, we find
that most of the benefits $\swish$ are realized by using them only in the deeper layers. Thus in our architectures we only use
$\hardswish$ at the second half of the model. We refer to the tables \ref{table:v3-large} and \ref{table:v3-small} for the precise
layout.

Even with these optimizations, $\hardswish$  still introduces some latency cost. However
as we demonstrate in section \ref{section:experiments} the net effect on accuracy and
latency is positive with no optimizations and substantial when using an optimized implementation based on a piece-wise function.

\subsection{Large squeeze-and-excite}
In ~\cite{mnasnet}, the size of the squeeze-and-excite bottleneck was relative the size of the convolutional bottleneck. Instead, we replace them all to fixed to be 1/4 of the number of channels in  expansion layer.  We find that doing so increases the accuracy, at the modest increase of number of  parameters, and no discernible latency cost. 

\subsection{MobileNetV3 Definitions}
MobileNetV3 is defined as two models: MobileNetV3-Large and MobileNetV3-Small. These models are targeted at high and low resource use cases respectively. The models are created through applying platform-aware NAS and NetAdapt for network search and incorporating the network improvements defined in this section. See table \ref{table:v3-large} and \ref{table:v3-small} for full specification of our networks.

\begin{table}[t]
\centering
\vspace{0pt}
\scalebox{0.8}{
\begin{tabular}{c|c|c|c|c|c|c}
\toprule[0.2em]
Input & Operator                   &  exp size& $\#out $ & SE& NL& $s$ \\
\toprule[0.2em]
$224^2\times3$  &    conv2d        &  -        &  16     & -           & HS  & 2\\
$112^2\times16$ &    bneck, 3x3    &  16       & 16      & -           & RE  & 1\\  %
$112^2\times16$ &    bneck, 3x3    &  64       & 24      & -           & RE  & 2\\   %
$56^2\times24$  &    bneck, 3x3   &  72       & 24       & -           & RE  & 1\\  %
$56^2\times24$  &    bneck, 5x5    &  72       & 40      &\checkmark  & RE  & 2 \\  %
$28^2\times40$  &    bneck, 5x5    &  120      & 40      &\checkmark  & RE  & 1 \\ %
$28^2\times40$  &    bneck, 5x5    &  120      & 40      &\checkmark  & RE  & 1 \\ %
$28^2\times40$  &    bneck, 3x3    &  240      & 80      & -           & HS  & 2 \\ %
$14^2\times80$  &    bneck, 3x3    &  200      & 80      & -           & HS  & 1 \\ %
$14^2\times80$  &    bneck, 3x3    &  184      & 80      & -           & HS  & 1 \\ %
$14^2\times80$  &    bneck, 3x3    &  184      & 80      & -           & HS  & 1 \\ %
$14^2\times80$  &    bneck, 3x3    &  480      & 112     &\checkmark  & HS  & 1 \\ %
$14^2\times112$ &    bneck, 3x3    &  672      & 112     &\checkmark  & HS  & 1 \\ %
$14^2\times112$ &    bneck, 5x5    &  672      & 160     &\checkmark  & HS  & 2 \\ %
$7^2\times160$ &    bneck, 5x5    &  960      & 160     &\checkmark  & HS  & 1 \\ %
$7^2\times160$  &    bneck, 5x5    &  960      & 160     &\checkmark  & HS  & 1  \\  %
$7^2\times160$  &    conv2d, 1x1   &  -        & 960     & -           & HS  & 1 \\ %
$7^2\times960$  &    pool, 7x7     &  -        & -       & -           & -  & 1 \\   
$1^2\times 960$ & conv2d 1x1, NBN   &  -        & 1280    & -           & HS & 1 \\
$1^2\times 1280$ & conv2d 1x1, NBN  &  -        & k       & -           &   -   & 1 \\
\toprule[0.2em]
\end{tabular}
}
\caption{Specification for MobileNetV3-Large. SE denotes whether there is a Squeeze-And-Excite in that block. NL denotes the type of nonlinearity used. 
Here, HS denotes $\hardswish$ and RE denotes $\relu$. NBN denotes no batch normalization.
$s$ denotes stride. }
\label{table:v3-large}
\end{table}

\begin{table}[t]
\centering
\vspace{0pt}
\scalebox{0.8}{
\begin{tabular}{c|c|c|c|c|c|c|c}
\toprule[0.2em]
Input & Operator                   &  exp size& $\#out $ & SE & NL  &$s$ \\
\toprule[0.2em]
$224^2\times3$  &    conv2d, 3x3   &  -         &  16   &  -  & HS  & 2\\
$112^2\times16$ &    bneck, 3x3    &  16        &  16   & \checkmark & RE  & 2\\  %
$56^2\times16$  &    bneck, 3x3    &  72        &  24   & -          & RE & 2\\  %
$28^2\times24$  &    bneck, 3x3    &  88        &  24   & -          & RE & 1 \\  %
$28^2\times24$  &    bneck, 5x5    &  96        & 40    & \checkmark & HS & 2 \\ %
$14^2\times40$  &    bneck, 5x5    &  240       & 40    & \checkmark & HS & 1 \\ %
$14^2\times40$  &    bneck, 5x5    &  240       & 40    & \checkmark & HS & 1 \\ %
$14^2\times40$  &    bneck, 5x5    &  120       & 48    & \checkmark & HS & 1 \\ %
$14^2\times48$  &    bneck, 5x5    &  144       & 48    & \checkmark & HS & 1 \\ %
$14^2\times48$ &    bneck, 5x5     &  288       & 96    & \checkmark & HS & 2 \\ %
$7^2\times96$  &    bneck, 5x5     &  576       & 96    & \checkmark & HS & 1 \\ %
$7^2\times96$   &    bneck, 5x5     &  576       & 96    & \checkmark & HS & 1 \\ %
$7^2\times96$   &    conv2d, 1x1    &  -         & 576   & \checkmark & HS & 1 \\ %
$7^2\times576$  &   pool, 7x7      &  -         & -     & -          & - & 1 \\   
$1^2\times 576$ & conv2d 1x1, NBN  &  -         & 1024  & -          & HS & 1 \\
$1^2\times 1024$& conv2d 1x1, NBN  &  -         & k     & -          & -  & 1 \\
\toprule[0.2em]
\end{tabular}
}
\caption{
Specification for MobileNetV3-Small. See table \ref{table:v3-large} for notation.
}
\label{table:v3-small}
\end{table}

\section{Experiments}
\label{section:experiments}
We present experimental results to demonstrate the effectiveness of the new MobileNetV3 models. We report results on classification, detection and segmentation. We also report various ablation studies to shed light on the effects of various design decisions.

\subsection{Classification}
As has become standard, we use ImageNet\cite{russakovsky:2015:ILS:2846547.2846559} for all our classification experiments and compare accuracy versus various measures of resource usage such as latency and multiply adds (MAdds).  
\subsubsection{Training setup}
We train our models using synchronous training setup  on
4x4 TPU Pod \cite{tpu-intro} using standard tensorflow RMSPropOptimizer with 0.9 momentum. We use the initial learning rate of 0.1, with batch size 4096 (128 images per chip), and learning rate decay rate of 0.01 every 3 epochs. We use dropout of 0.8, and l2 weight decay 1e-5 and the same image preprocessing as Inception~\cite{InceptionV4}. Finally we use exponential moving average with decay 0.9999. All our convolutional layers use batch-normalization layers with average decay of 0.99. 

\subsubsection{Measurement setup}
To measure latencies we use standard Google Pixel phones and run all networks through the standard TFLite
Benchmark Tool. We use single-threaded large core in all our measurements. We don't report multi-core inference
time, since we find this setup not very practical for mobile applications. We contributed an atomic h-swish
operator to tensorflow lite, and it is now default in the latest version. We show the impact of optimized 
h-swish on figure \ref{fig:v3_progression}.

\subsection{Results}
As can be seen on figure \ref{fig:latency_vs_top_1} our models outperform the current state of the art such as
MnasNet \cite{mnasnet}, ProxylessNas \cite{proxyless}
and MobileNetV2 \cite{mobilenetv2}. We report the floating point performance on different Pixel phones in the
table \ref{table:float-latencies}. We include quantization results in table \ref{table:quantized-latencies}.

\begin{figure}[t!]
    \centering
    \includegraphics[width=0.47\textwidth]{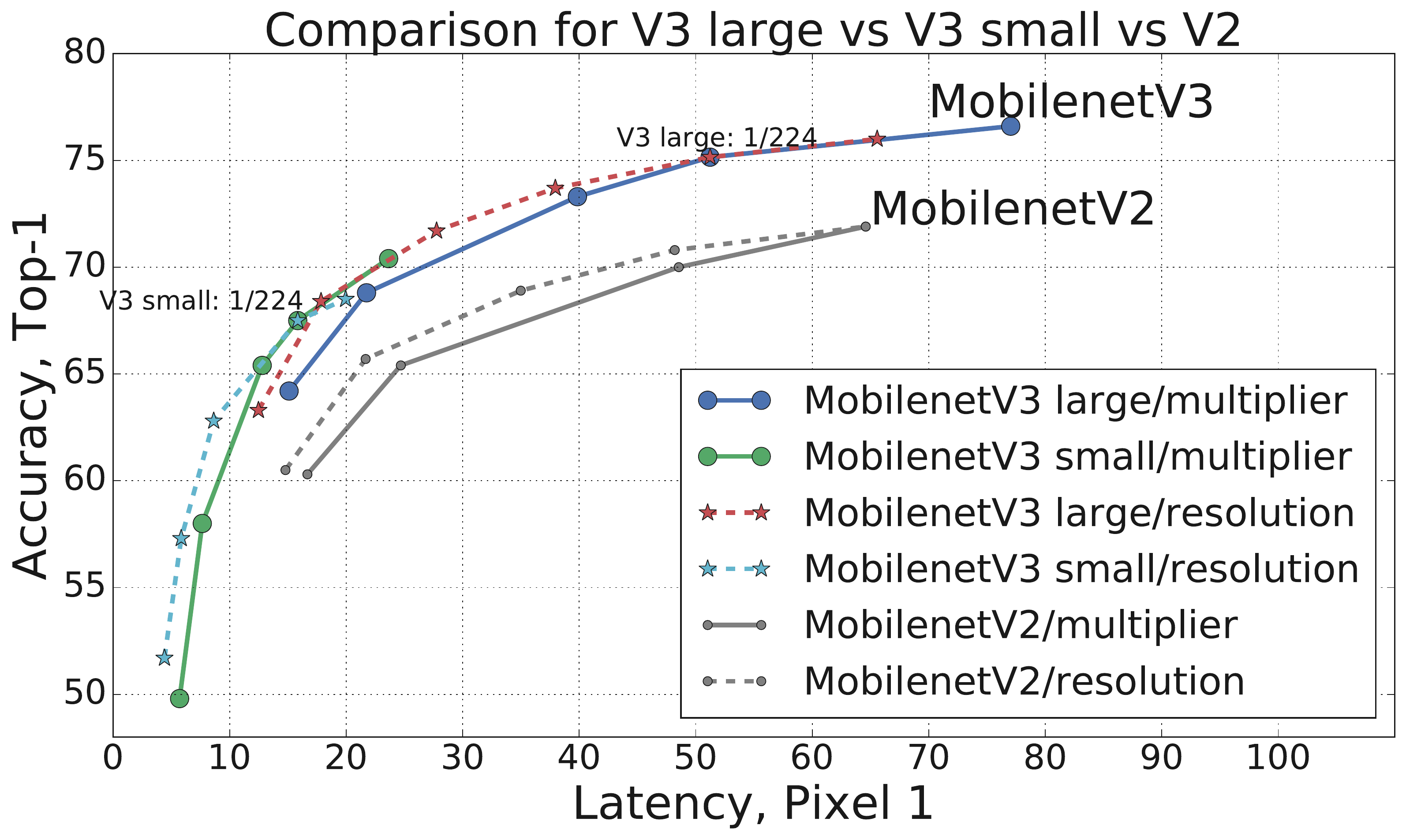}
    \caption{Performance of MobileNetV3 as a function of different multipliers and resolutions. In our experiments we have used
    multipliers 0.35, 0.5, 0.75, 1.0 and 1.25, with a fixed resolution of 224, 
    and resolutions 96, 128, 160, 192, 224 and 256 with a fixed depth multiplier of 1.0. Best viewed in color. Top-1 accuracy is on ImageNet and latency is in ms.}
    \label{fig:resolution-vs-latency}
\end{figure}

In figure \ref{fig:resolution-vs-latency} we show the MobileNetV3 performance trade-offs as a function of
multiplier and resolution.
Note how MobileNetV3-Small outperforms the MobileNetV3-Large with multiplier scaled to match the performance by
nearly 3\%. On the
other hand, resolution provides an even better trade-offs than multiplier. However, it should be noted that resolution is often determined by the 
problem (e.g. segmentation and detection problem generally require higher resolution), and thus can't always be used as a tunable parameter.

\begin{table}[t]
\centering
\vspace{0pt}
\scalebox{0.85}{
\begin{tabular}{l|c|c|c|c|c|c}
\toprule[0.2em]
Network                  & Top-1& MAdds &Params   & P-1 & P-2 & P-3 \\
\hline
V3-Large 1.0              &\textbf{75.2} & \textbf{219} & 5.4M  & \textbf{51} & \textbf{61}   & \textbf{44} \\
V3-Large 0.75            & 73.3 & 155 & 4.0M      & 39     & 46   & 40 \\
MnasNet-A1               & 75.2 & 315 & 3.9M    & 71     & 86   & 61   \\
Proxyless\cite{proxyless}& 74.6 & 320 & 4.0M      & 72     & 84   & 60   \\
V2 1.0                   & 72.0 & 300 & 3.4M    & 64     & 76   & 56 \\
\hline
V3-Small 1.0             & \textbf{67.4} &  \textbf{56}   & 2.5M  & \textbf{15.8}    & \textbf{19.4}     & \textbf{14.4}\\
V3-Small 0.75            & 65.4 &  44   & 2.0M  & 12.8    & 15.6     & 11.7\\
Mnas-small \cite{mnasnet}   & 64.9  & 65.1 & 1.9M  & 20.3    & 24.2     & 17.2 \\ 
V2 0.35                  & 60.8 &  59.2 & 1.6M  & 16.6    & 19.6     & 13.9 \\

\hline
\toprule[0.2em]
\end{tabular}
}
\caption{Floating point performance on the Pixel family of phones (P-$n$ denotes a Pixel-$n$ phone). All latencies are in ms and are measured using a single large core with a batch size of one. Top-1 accuracy is on ImageNet.}
\label{table:float-latencies}
\end{table}

\begin{table}[t]
\centering
\vspace{0pt}
\begin{tabular}{l|c|c|c|c}
\toprule[0.2em]
Network        & Top-1 & P-1 & P-2 & P-3 \\
\hline
\textbf{V3-Large 1.0} & \textbf{73.8}   & 44     &  42.5 & 31.7\\ 
 V2 1.0                & 70.9           & 52     &  48.3 & 37.0  \\
\hline
 \textbf{V3-Small}     & \textbf{64.9}  & 15.5   & 14.9 & 10.7\\
 V2 0.35               & 57.2           & 16.7   & 15.6 & 11.9 \\
\hline
\toprule[0.2em]
\end{tabular}
\caption{Quantized performance. All latencies are in ms. The inference latency is measured using a single large core on the respective Pixel 1/2/3 device.}
\label{table:quantized-latencies}
\end{table}

\subsubsection{Ablation study}
\paragraph{Impact of non-linearities} In table \ref{table:nonlinearities} we study the choice of where to insert $\hardswish$ nonlinearities as well as the improvements of using an optimized implementation over a naive implementation. It can be seen that using an optimized implementation of $\hardswish$ saves 6ms (more than $10\%$ of the runtime). Optimized $\hardswish$ only adds an additional 1ms compared to traditional ReLU. 

\begin{figure}
    \centering
     \includegraphics[width=0.47\textwidth]{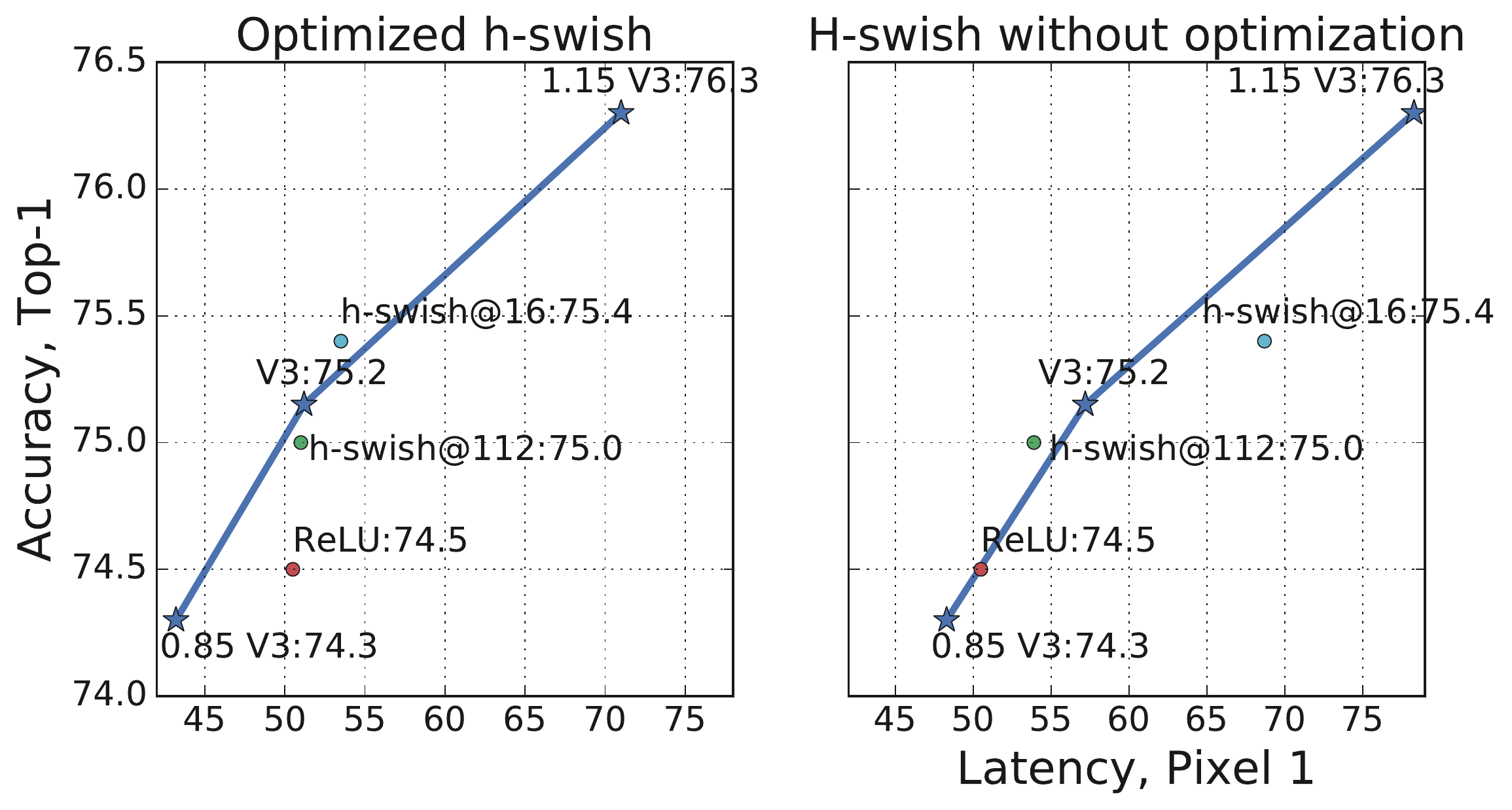}
    \caption{Impact of h-swish vs ReLU on latency for optimized and non-optimized h-swish. The curve shows a frontier of using depth multiplier. Note that placing $\hardswish$  at all layers with 80 channels or more (V3) provides the best trade-offs  for both optimized h-swish and non-optimized h-swish.  Top-1 accuracy is on ImageNet and latency is in ms. }
    \label{fig:nonlinearities}
\end{figure}

Figure \ref{fig:nonlinearities} 
shows the efficient frontier based on nonlinearity choices and network width. MobileNetV3 uses h-swish in the middle of the network and clearly dominates ReLU. It is interesting to note that adding $\hardswish$ to the entire network is slightly better than the interpolated frontier of widening the network.

\begin{table}[t]
\centering
\begin{tabular}{l|c|c|c}
\toprule[0.2em]
         & Top-1  &  P-1 & P-1 (no-opt) \\
\hline
V3-Large 1.0       & 75.2   & 51.4 & 57.5 \\ %
\hline
$\relu$  & 74.5 ({\color{red}-.7\%})   & 50.5 ({\color{green}-1\%}) & 50.5 \\
$\hardswish@16$      & 75.4 (+.2\%)   & 53.5 ({\color{red}+4\%}) & 68.9 \\
$\hardswish@112$ & 75.0 (-.3\%)  & 51  ({-0.5\%}) & 54.4  \\
\toprule[0.2em]
\end{tabular}
\caption{Effect of non-linearities on MobileNetV3-Large. In $\hardswish@N$, $N$ denotes the number of channels, in the first layer that has $\hardswish$ enabled.  The third column shows the runtime without optimized h-swish. Top-1 accuracy is on ImageNet and latency is in ms.%
}
\label{table:nonlinearities}
\end{table}

\begin{figure}
    \centering
    \includegraphics[width=0.47\textwidth]{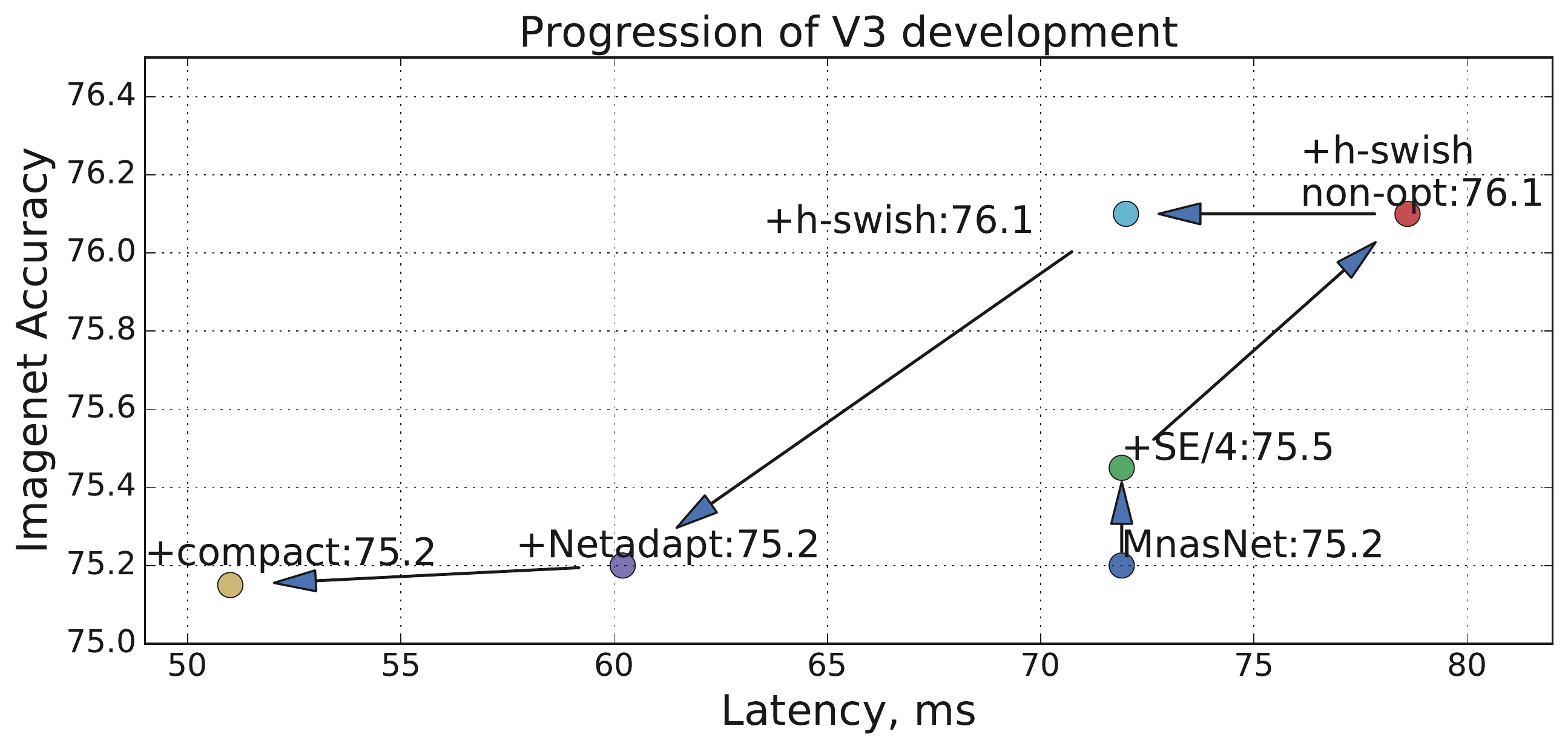}
    \caption{Impact of individual components in the development of MobileNetV3. Progress is measured by moving up and to the left.}
    \label{fig:v3_progression}
\end{figure}

\paragraph{Impact of other components} In figure \ref{fig:v3_progression}
we show how introduction of different components moved along the latency/accuracy curve. 

\subsection{Detection}
We use MobileNetV3 as a drop-in replacement for the backbone feature extractor in SSDLite~\cite{mobilenetv2} and compare with other backbone networks on COCO dataset~\cite{COCO}. 

Following MobileNetV2~\cite{mobilenetv2}, we attach the first layer of SSDLite to the last feature extractor layer that has an output stride of $16$, and attach the second layer of SSDLite to the last feature extractor layer that has an output stride of $32$. Following the detection literature, we refer to these two feature extractor layers as $C4$ and $C5$, respectively. For MobileNetV3-Large, $C4$ is the expansion layer of the $13$-th bottleneck block. For MobileNetV3-Small, $C4$ is the expansion layer of the $9$-th bottleneck block. For both networks, $C5$ is the layer immediately before pooling.

We additionally reduce the channel counts of all feature layers between $C4$ and $C5$ by $2$. This is because the last few layers of MobileNetV3 are tuned to output $1000$ classes, which may be redundant when transferred to COCO with $90$ classes. 

The results on COCO test set are given in Tab.~\ref{tab:detection}. With the channel reduction, MobileNetV3-Large is $27\%$ faster than MobileNetV2 with near identical mAP. MobileNetV3-Small with channel reduction is also $2.4$ and $0.5$ mAP higher than MobileNetV2 and MnasNet while being $35\%$ faster. For both MobileNetV3 models the channel reduction trick contributes to approximately $15\%$ latency reduction with no mAP loss, suggesting that Imagenet classification and COCO object detection may prefer different feature extractor shapes.

\begin{table}[t]
\centering
\scalebox{0.84}{
\begin{tabular}{l|c | c| cc}
\toprule[0.2em]
Backbone & mAP & Latency (ms) & Params (M) & MAdds (B) \\ 
\midrule
V1      & 22.2 & 228 & 5.1 & 1.3 \\
V2      & 22.1 & 162 & 4.3 & 0.80 \\
MnasNet & 23.0 & 174 & 4.88 & 0.84 \\
\midrule
V3      & 22.0 & 137 & 4.97 & 0.62    \\
{\bf V3$^\dagger$ }     & 22.0 & 119 & 3.22 & 0.51   \\
\midrule
\midrule
V2 $0.35$ & 13.7 & 66 & 0.93 & 0.16 \\
V2 $0.5$ & 16.6 & 79 & 1.54 & 0.27 \\
MnasNet $0.35$ & 15.6 & 68 & 1.02 & 0.18 \\
MnasNet $0.5$ & 18.5 & 85 & 1.68 & 0.29 \\
\midrule
V3-Small & 16.0 & 52 & 2.49 & 0.21 \\
{\bf V3-Small$^\dagger$} & 16.1 & 43 & 1.77 & 0.16 \\
\bottomrule
\end{tabular}
}
\caption{Object detection results of SSDLite with different backbones on COCO test set. $^\dagger$: Channels in the blocks between $C4$ and $C5$ are reduced by a factor of 2.}
\label{tab:detection}
\end{table}

\subsection{Semantic Segmentation}

\begin{figure}[!t]
    \centering
    \includegraphics[width=0.47\textwidth]{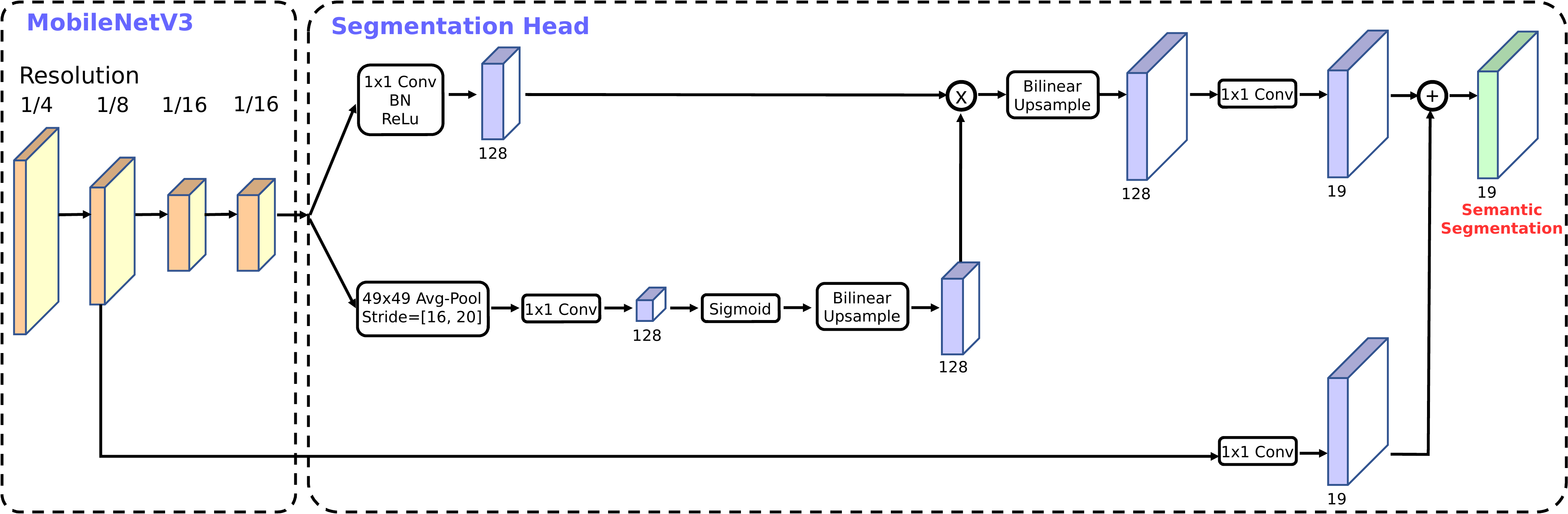}
    \caption{Building on MobileNetV3, the proposed segmentation head, Lite R-ASPP, delivers fast semantic segmentation results while mixing features from multiple resolutions.}
    \label{fig:mobilenetv3_seg}
\end{figure}

In this subsection, we employ MobileNetV2 \cite{mobilenetv2} and the proposed MobileNetV3 as network backbones for the task of {\it mobile} semantic segmentation. Additionally, we compare two segmentation heads. The first one, referred to as R-ASPP, was proposed in \cite{mobilenetv2}. R-ASPP is a {\it reduced} design of the Atrous Spatial Pyramid Pooling module \cite{DeepLabV2, DeepLabV3,DeepLabV3Plus}, which adopts only two branches consisting of a $1 \times 1$ convolution and a global-average pooling operation \cite{ParseNet,zhao2017pyramid}. In this work, we propose another light-weight segmentation head, referred to as Lite R-ASPP (or LR-ASPP), as shown in Fig.~\ref{fig:mobilenetv3_seg}. Lite R-ASPP, improving over R-ASPP, deploys the global-average pooling in a fashion similar to the Squeeze-and-Excitation module \cite{squeezeandexcite}, in which we employ a large pooling kernel with a large stride (to save some computation) and only one $1\times1$ convolution in the module. We apply atrous convolution \cite{Holschneider1989real,Sermanet2013Overfeat,Papandreou2014untangling,DeepLabV1} to the last block of MobileNetV3 to extract denser features, and further add a skip connection \cite{long2014fully} from low-level features to capture more detailed information.

We conduct the experiments on the Cityscapes dataset \cite{Cordts2016Cityscapes} with metric mIOU \cite{PASCAL}, and only exploit the `fine' annotations. We employ the same training protocol as \cite{DeepLabV3, mobilenetv2}. All our models are trained from scratch without pretraining on ImageNet \cite{russakovsky:2015:ILS:2846547.2846559}, and are evaluated with a {\it single-scale} input. Similar to object detection, we observe that we could reduce the channels in the last block of network backbone by a factor of 2 without degrading the performance significantly. We think it is because the backbone is designed for 1000 classes ImageNet image classification \cite{russakovsky:2015:ILS:2846547.2846559} while there are only 19 classes on Cityscapes, implying there is some channel redundancy in the backbone.

We report our Cityscapes validation set results in Tab.~\ref{tab:cityscapes_val}. As shown in the table, we observe that (1) reducing the channels in the last block of network backbone by a factor of 2 significantly improves the speed while maintaining similar performances (row 1 vs. row 2, and row 5 vs. row 6), (2) the proposed segmentation head LR-ASPP is slightly faster than R-ASPP \cite{mobilenetv2} while performance is improved (row 2 vs. row 3, and row 6 vs. row 7), (3) reducing the filters in the segmentation head from 256 to 128 improves the speed at the cost of slightly worse performance (row 3 vs. row 4, and row 7 vs. row 8), (4) when employing the same setting, MobileNetV3 model variants attain similar performance while being slightly faster than MobileNetV2 counterparts (row 1 vs. row 5, row 2 vs. row 6, row 3 vs. row 7, and row 4 vs. row 8), (5) MobileNetV3-Small attains similar performance as MobileNetV2-0.5 while being faster, and (6) MobileNetV3-Small is significantly better than MobileNetV2-0.35 while yielding similar speed.

Tab.~\ref{tab:cityscapes_test} shows our Cityscapes test set results. Our segmentation models with MobileNetV3 as network backbone outperforms ESPNetv2 \cite{DBLP:journals/corr/abs-1811-11431}, CCC2 \cite{DBLP:journals/corr/abs-1812-04920}, and ESPNetv1 \cite{DBLP:journals/corr/abs-1811-11431} by 6.4\%, 10.6\%, 12.3\%, respectively while being faster in terms of MAdds. The performance drops slightly by 0.6\% when not employing the atrous convolution to extract dense feature maps in the last block of MobileNetV3, but the speed is improved to 1.98B (for half-resolution inputs), which is 1.36, 1.59, and 2.27 times faster than ESPNetv2, CCC2, and ESPNetv1, respectively. Furthermore, our models with MobileNetV3-Small as network backbone still outperforms all of them by at least a healthy margin of 2.1\%. %

\begin{table}[t]
\centering
\scalebox{0.64}{
\begin{tabular}{l l|c c c|c c c c c}
\toprule[0.2em]
N & Backbone & RF2 & SH & F & mIOU & Params & MAdds & CPU (f) & CPU (h) \\
\midrule
1 & V2                & - &  $\times$ & 256 & 72.84 & 2.11M & 21.29B & 3.90s & 1.02s \\
2 & V2                & \checkmark &  $\times$ & 256 & 72.56 & 1.15M & 13.68B & 3.03s & 793ms \\
3 & V2                & \checkmark & \checkmark & 256 & 72.97 & 1.02M & 12.83B & 2.98s & 786ms \\
4 & V2                & \checkmark & \checkmark & 128 & 72.74 & 0.98M & 12.57B & 2.89s & 766ms \\
\midrule
5 & V3                & - &  $\times$ & 256 & 72.64 & 3.60M & 18.43B & 3.55s & 906ms \\
6 & V3                & \checkmark &  $\times$ & 256 & 71.91  & 1.76M & 11.24B & 2.60s & 668ms \\
7 & V3                & \checkmark & \checkmark & 256 & 72.37 & 1.63M & 10.33B & 2.55s & 659ms \\
8 & V3                & \checkmark & \checkmark & 128 & 72.36 & 1.51M &  9.74B & 2.47s & 657ms \\
\midrule \midrule
9 & V2 0.5            & \checkmark & \checkmark & 128 & 68.57 & 0.28M & 4.00B & 1.59s & 415ms \\
10 & V2 0.35           & \checkmark & \checkmark & 128 & 66.83 & 0.16M & 2.54B & 1.27s & 354ms \\
\midrule
11 & V3-Small & \checkmark & \checkmark & 128 & 68.38 & 0.47M &  2.90B & 1.21s & 327ms \\
\bottomrule
\end{tabular}
}
\caption{Semantic segmentation results on Cityscapes {\it val} set. {\bf RF2}: {\bf R}educe the {\bf F}ilters in the last block by a factor of {\bf 2}. V2 0.5 and V2 0.35 are MobileNetV2 with depth multiplier = 0.5 and 0.35, respectively. {\bf SH:} {\bf S}egmentation {\bf H}ead, where $\times$ employs the R-ASPP while $\checkmark$ employs the proposed LR-ASPP. {\bf F:} Number of {\bf F}ilters used in the Segmentation Head. {\bf CPU (f):} CPU time measured on a single large core of Pixel 3 (floating point) w.r.t. a {\bf full-resolution} input (i.e., $1024\times2048$). {\bf CPU (h):} CPU time measured w.r.t. a {\bf half-resolution} input (i.e., $512\times1024$). Row 8, and 11 are our MobileNetV3 segmentation candidates.}
\label{tab:cityscapes_val}
\end{table}

\begin{table}[t]
\centering
\scalebox{0.68}{
\begin{tabular}{l|c | c c c c c}
\toprule[0.2em]
Backbone & OS & mIOU & MAdds (f) & MAdds (h) & CPU (f) & CPU (h) \\ 
\midrule
V3          & 16 & 72.6 & 9.74B & 2.48B & 2.47s & 657ms \\
V3          & 32 & 72.0 & 7.74B & 1.98B & 2.06s & 534ms \\
\midrule
V3-Small    & 16 & 69.4 & 2.90B & 0.74B & 1.21s & 327ms \\
V3-Small    & 32 & 68.3 & 2.06B & 0.53B & 1.03s & 275ms \\
\midrule \midrule
ESPNetv2 \cite{DBLP:journals/corr/abs-1811-11431} & - & 66.2 & - & 2.7B  & - & -\\
CCC2 \cite{DBLP:journals/corr/abs-1812-04920} & - & 62.0 & - & 3.15B & - & - \\
ESPNetv1 \cite{DBLP:conf/eccv/MehtaRCSH18} & - & 60.3 & - & 4.5B & - & - \\
\bottomrule
\end{tabular}
}
\caption{Semantic segmentation results on Cityscapes test set. {\bf OS:} {\bf O}utput {\bf S}tride, the ratio of input image spatial resolution to backbone output resolution. When OS = 16, atrous convolution is applied in the last block of backbone. When OS = 32, no atrous convolution is used. {\bf MAdds (f):} Multiply-Adds measured w.r.t. a {\bf full-resolution} input (i.e., $1024\times2048$). {\bf MAdds (h):} Multiply-Adds measured w.r.t. a {\bf half-resolution} input (i.e., $512\times1024$). {\bf CPU (f):} CPU time measured on a single large core of Pixel 3 (floating point) w.r.t. a {\bf full-resolution} input (i.e., $1024\times2048$). {\bf CPU (h):} CPU time measured w.r.t. a {\bf half-resolution} input (i.e., $512\times1024$). ESPNet \cite{DBLP:conf/eccv/MehtaRCSH18,DBLP:journals/corr/abs-1811-11431} and CCC2 \cite{DBLP:journals/corr/abs-1812-04920} take half resolution inputs, while our models directly take full-resolution inputs.}
\label{tab:cityscapes_test}
\end{table}
 
\section{Conclusions and future work}
\label{section:conclusions}

In this paper we introduced MobileNetV3 Large and Small models demonstrating new state of the art in mobile classification, detection and segmentation. We have described our efforts to harness multiple network architecture search algorithms as well as advances in network design to deliver the next generation of mobile models. We have also shown how to adapt nonlinearities like swish and apply squeeze and excite in a quantization friendly and efficient manner introducing them into the mobile model domain as effective tools. We also introduced a new form of lightweight segmentation decoders called LR-ASPP. While it remains an open question of how best to blend automatic search techniques with human intuition, we are pleased to present these first positive results and will continue to refine methods as future work.

{\bf Acknowledgements:} We would like to thank Andrey Zhmoginov, Dmitry Kalenichenko, Menglong Zhu, Jon Shlens, Xiao Zhang, Benoit Jacob, Alex Stark, Achille Brighton and Sergey Ioffe for helpful feedback and discussion.

{\small
\bibliographystyle{ieee_fullname}
\bibliography{egbib}

\begin{thebibliography}{10}\itemsep=-1pt

\bibitem{tensorflow2015-whitepaper}
Mart\'{\i}n Abadi, Ashish Agarwal, Paul Barham, Eugene Brevdo, Zhifeng Chen,
  Craig Citro, Greg~S. Corrado, Andy Davis, Jeffrey Dean, Matthieu Devin,
  Sanjay Ghemawat, Ian Goodfellow, Andrew Harp, Geoffrey Irving, Michael Isard,
  Yangqing Jia, Rafal Jozefowicz, Lukasz Kaiser, Manjunath Kudlur, Josh
  Levenberg, Dan Man\'{e}, Rajat Monga, Sherry Moore, Derek Murray, Chris Olah,
  Mike Schuster, Jonathon Shlens, Benoit Steiner, Ilya Sutskever, Kunal Talwar,
  Paul Tucker, Vincent Vanhoucke, Vijay Vasudevan, Fernanda Vi\'{e}gas, Oriol
  Vinyals, Pete Warden, Martin Wattenberg, Martin Wicke, Yuan Yu, and Xiaoqiang
  Zheng.
\newblock {TensorFlow}: Large-scale machine learning on heterogeneous systems,
  2015.
\newblock Software available from tensorflow.org.

\bibitem{hard-swish}
R. Avenash and P. Vishawanth.
\newblock Semantic segmentation of satellite images using a modified cnn with
  hard-swish activation function.
\newblock In {\em VISIGRAPP}, 2019.

\bibitem{BakerGNR16}
Bowen Baker, Otkrist Gupta, Nikhil Naik, and Ramesh Raskar.
\newblock Designing neural network architectures using reinforcement learning.
\newblock {\em CoRR}, abs/1611.02167, 2016.

\bibitem{ModelCompression}
Cristian Bucilu\v{a}, Rich Caruana, and Alexandru Niculescu-Mizil.
\newblock Model compression.
\newblock In {\em Proceedings of the 12th ACM SIGKDD International Conference
  on Knowledge Discovery and Data Mining}, KDD '06, pages 535--541, New York,
  NY, USA, 2006. ACM.

\bibitem{proxyless}
Han Cai, Ligeng Zhu, and Song Han.
\newblock Proxylessnas: Direct neural architecture search on target task and
  hardware.
\newblock {\em CoRR}, abs/1812.00332, 2018.

\bibitem{DeepLabV1}
Liang-Chieh Chen, George Papandreou, Iasonas Kokkinos, Kevin Murphy, and Alan~L
  Yuille.
\newblock Semantic image segmentation with deep convolutional nets and fully
  connected crfs.
\newblock In {\em ICLR}, 2015.

\bibitem{DeepLabV2}
Liang-Chieh Chen, George Papandreou, Iasonas Kokkinos, Kevin Murphy, and Alan~L
  Yuille.
\newblock Deeplab: Semantic image segmentation with deep convolutional nets,
  atrous convolution, and fully connected crfs.
\newblock {\em TPAMI}, 2017.

\bibitem{DeepLabV3}
Liang-Chieh Chen, George Papandreou, Florian Schroff, and Hartwig Adam.
\newblock Rethinking atrous convolution for semantic image segmentation.
\newblock {\em CoRR}, abs/1706.05587, 2017.

\bibitem{DeepLabV3Plus}
Liang-Chieh Chen, Yukun Zhu, George Papandreou, Florian Schroff, and Hartwig
  Adam.
\newblock Encoder-decoder with atrous separable convolution for semantic image
  segmentation.
\newblock In {\em ECCV}, 2018.

\bibitem{Cordts2016Cityscapes}
Marius Cordts, Mohamed Omran, Sebastian Ramos, Timo Rehfeld, Markus Enzweiler,
  Rodrigo Benenson, Uwe Franke, Stefan Roth, and Bernt Schiele.
\newblock The cityscapes dataset for semantic urban scene understanding.
\newblock In {\em CVPR}, 2016.

\bibitem{binaryconnect}
Matthieu Courbariaux, Yoshua Bengio, and Jean{-}Pierre David.
\newblock Binaryconnect: Training deep neural networks with binary weights
  during propagations.
\newblock {\em CoRR}, abs/1511.00363, 2015.

\bibitem{ChamNet}
Xiaoliang Dai, Peizhao Zhang, Bichen Wu, Hongxu Yin, Fei Sun, Yanghan Wang,
  Marat Dukhan, Yunqing Hu, Yiming Wu, Yangqing Jia, Peter Vajda, Matt
  Uyttendaele, and Niraj~K. Jha.
\newblock Chamnet: Towards efficient network design through platform-aware
  model adaptation.
\newblock {\em CoRR}, abs/1812.08934, 2018.

\bibitem{silu}
Stefan Elfwing, Eiji Uchibe, and Kenji Doya.
\newblock Sigmoid-weighted linear units for neural network function
  approximation in reinforcement learning.
\newblock {\em CoRR}, abs/1702.03118, 2017.

\bibitem{PASCAL}
Mark Everingham, S.~M.~Ali Eslami, Luc~Van Gool, Christopher K.~I. Williams,
  John Winn, and Andrew Zisserma.
\newblock The pascal visual object classes challenge – a retrospective.
\newblock {\em IJCV}, 2014.

\bibitem{AMC}
Yihui He and Song Han.
\newblock {AMC:} automated deep compression and acceleration with reinforcement
  learning.
\newblock In {\em ECCV}, 2018.

\bibitem{gelu}
Dan Hendrycks and Kevin Gimpel.
\newblock Bridging nonlinearities and stochastic regularizers with gaussian
  error linear units.
\newblock {\em CoRR}, abs/1606.08415, 2016.

\bibitem{Distill}
Geoffrey Hinton, Oriol Vinyals, and Jeffrey Dean.
\newblock Distilling the knowledge in a neural network.
\newblock In {\em NIPS Deep Learning and Representation Learning Workshop},
  2015.

\bibitem{Holschneider1989real}
Matthias Holschneider, Richard Kronland-Martinet, Jean Morlet, and Ph
  Tchamitchian.
\newblock A real-time algorithm for signal analysis with the help of the
  wavelet transform.
\newblock In {\em Wavelets: Time-Frequency Methods and Phase Space}, pages
  289--297. Springer Berlin Heidelberg, 1989.

\bibitem{MobilenetV1}
Andrew~G. Howard, Menglong Zhu, Bo Chen, Dmitry Kalenichenko, Weijun Wang,
  Tobias Weyand, Marco Andreetto, and Hartwig Adam.
\newblock Mobilenets: Efficient convolutional neural networks for mobile vision
  applications.
\newblock {\em CoRR}, abs/1704.04861, 2017.

\bibitem{squeezeandexcite}
J. {Hu}, L. {Shen}, and G. {Sun}.
\newblock {Squeeze-and-Excitation Networks}.
\newblock {\em ArXiv e-prints}, Sept. 2017.

\bibitem{CondenseNet}
Gao Huang, Shichen Liu, Laurens van~der Maaten, and Kilian~Q. Weinberger.
\newblock Condensenet: An efficient densenet using learned group convolutions.
\newblock {\em CoRR}, abs/1711.09224, 2017.

\bibitem{SqueezeNet}
Forrest~N. Iandola, Matthew~W. Moskewicz, Khalid Ashraf, Song Han, William~J.
  Dally, and Kurt Keutzer.
\newblock Squeezenet: Alexnet-level accuracy with 50x fewer parameters and
  {\textless}1mb model size.
\newblock {\em CoRR}, abs/1602.07360, 2016.

\bibitem{fakequant}
Benoit Jacob, Skirmantas Kligys, Bo Chen, Menglong Zhu, Matthew Tang, Andrew
  Howard, Hartwig Adam, and Dmitry Kalenichenko.
\newblock Quantization and training of neural networks for efficient
  integer-arithmetic-only inference.
\newblock In {\em The IEEE Conference on Computer Vision and Pattern
  Recognition (CVPR)}, June 2018.

\bibitem{tpu-intro}
Norman~P. Jouppi, Cliff Young, Nishant Patil, David~A. Patterson, Gaurav
  Agrawal, Raminder Bajwa, Sarah Bates, Suresh Bhatia, Nan Boden, Al Borchers,
  Rick Boyle, Pierre{-}luc Cantin, Clifford Chao, Chris Clark, Jeremy Coriell,
  Mike Daley, Matt Dau, Jeffrey Dean, Ben Gelb, Tara~Vazir Ghaemmaghami,
  Rajendra Gottipati, William Gulland, Robert Hagmann, Richard~C. Ho, Doug
  Hogberg, John Hu, Robert Hundt, Dan Hurt, Julian Ibarz, Aaron Jaffey, Alek
  Jaworski, Alexander Kaplan, Harshit Khaitan, Andy Koch, Naveen Kumar, Steve
  Lacy, James Laudon, James Law, Diemthu Le, Chris Leary, Zhuyuan Liu, Kyle
  Lucke, Alan Lundin, Gordon MacKean, Adriana Maggiore, Maire Mahony, Kieran
  Miller, Rahul Nagarajan, Ravi Narayanaswami, Ray Ni, Kathy Nix, Thomas
  Norrie, Mark Omernick, Narayana Penukonda, Andy Phelps, Jonathan Ross, Amir
  Salek, Emad Samadiani, Chris Severn, Gregory Sizikov, Matthew Snelham, Jed
  Souter, Dan Steinberg, Andy Swing, Mercedes Tan, Gregory Thorson, Bo Tian,
  Horia Toma, Erick Tuttle, Vijay Vasudevan, Richard Walter, Walter Wang, Eric
  Wilcox, and Doe~Hyun Yoon.
\newblock In-datacenter performance analysis of a tensor processing unit.
\newblock {\em CoRR}, abs/1704.04760, 2017.

\bibitem{quantwhite}
Raghuraman Krishnamoorthi.
\newblock Quantizing deep convolutional networks for efficient inference: {A}
  whitepaper.
\newblock {\em CoRR}, abs/1806.08342, 2018.

\bibitem{COCO}
Tsung-Yi Lin, Michael Maire, Serge Belongie, James Hays, Pietro Perona, Deva
  Ramanan, Piotr Doll{\'a}r, and C~Lawrence Zitnick.
\newblock Microsoft {COCO}: Common objects in context.
\newblock In {\em ECCV}, 2014.

\bibitem{ProgressiveNAS}
Chenxi Liu, Barret Zoph, Jonathon Shlens, Wei Hua, Li{-}Jia Li, Li Fei{-}Fei,
  Alan~L. Yuille, Jonathan Huang, and Kevin Murphy.
\newblock Progressive neural architecture search.
\newblock {\em CoRR}, abs/1712.00559, 2017.

\bibitem{DARTS}
Hanxiao Liu, Karen Simonyan, and Yiming Yang.
\newblock {DARTS:} differentiable architecture search.
\newblock {\em CoRR}, abs/1806.09055, 2018.

\bibitem{ParseNet}
Wei Liu, Andrew Rabinovich, and Alexander~C. Berg.
\newblock Parsenet: Looking wider to see better.
\newblock {\em CoRR}, abs/1506.04579, 2015.

\bibitem{long2014fully}
Jonathan Long, Evan Shelhamer, and Trevor Darrell.
\newblock Fully convolutional networks for semantic segmentation.
\newblock In {\em CVPR}, 2015.

\bibitem{DBLP:conf/eccv/MehtaRCSH18}
Sachin Mehta, Mohammad Rastegari, Anat Caspi, Linda~G. Shapiro, and Hannaneh
  Hajishirzi.
\newblock Espnet: Efficient spatial pyramid of dilated convolutions for
  semantic segmentation.
\newblock In {\em Computer Vision - {ECCV} 2018 - 15th European Conference,
  Munich, Germany, September 8-14, 2018, Proceedings, Part {X}}, pages
  561--580, 2018.

\bibitem{DBLP:journals/corr/abs-1811-11431}
Sachin Mehta, Mohammad Rastegari, Linda~G. Shapiro, and Hannaneh Hajishirzi.
\newblock Espnetv2: {A} light-weight, power efficient, and general purpose
  convolutional neural network.
\newblock {\em CoRR}, abs/1811.11431, 2018.

\bibitem{Papandreou2014untangling}
George Papandreou, Iasonas Kokkinos, and Pierre-Andre Savalle.
\newblock Modeling local and global deformations in deep learning: Epitomic
  convolution, multiple instance learning, and sliding window detection.
\newblock In {\em CVPR}, 2015.

\bibitem{DBLP:journals/corr/abs-1812-04920}
Hyojin Park, Youngjoon Yoo, Geonseok Seo, Dongyoon Han, Sangdoo Yun, and Nojun
  Kwak.
\newblock Concentrated-comprehensive convolutions for lightweight semantic
  segmentation.
\newblock {\em CoRR}, abs/1812.04920, 2018.

\bibitem{ENAS}
Hieu Pham, Melody~Y. Guan, Barret Zoph, Quoc~V. Le, and Jeff Dean.
\newblock Efficient neural architecture search via parameter sharing.
\newblock {\em CoRR}, abs/1802.03268, 2018.

\bibitem{swish}
Prajit Ramachandran, Barret Zoph, and Quoc~V. Le.
\newblock Searching for activation functions.
\newblock {\em CoRR}, abs/1710.05941, 2017.

\bibitem{Xnor-Net}
Mohammad Rastegari, Vicente Ordonez, Joseph Redmon, and Ali Farhadi.
\newblock Xnor-net: Imagenet classification using binary convolutional neural
  networks.
\newblock {\em CoRR}, abs/1603.05279, 2016.

\bibitem{russakovsky:2015:ILS:2846547.2846559}
Olga Russakovsky, Jia Deng, Hao Su, Jonathan Krause, Sanjeev Satheesh, Sean Ma,
  Zhiheng Huang, Andrej Karpathy, Aditya Khosla, Michael Bernstein,
  Alexander~C. Berg, and Li Fei-Fei.
\newblock Imagenet large scale visual recognition challenge.
\newblock {\em Int. J. Comput. Vision}, 115(3):211--252, Dec. 2015.

\bibitem{mobilenetv2}
Mark Sandler, Andrew~G. Howard, Menglong Zhu, Andrey Zhmoginov, and
  Liang{-}Chieh Chen.
\newblock Mobilenetv2: Inverted residuals and linear bottlenecks. mobile
  networks for classification, detection and segmentation.
\newblock {\em CoRR}, abs/1801.04381, 2018.

\bibitem{Sermanet2013Overfeat}
Pierre Sermanet, David Eigen, Xiang Zhang, Micha{\"e}l Mathieu, Rob Fergus, and
  Yann LeCun.
\newblock Overfeat: Integrated recognition, localization and detection using
  convolutional networks.
\newblock {\em arXiv:1312.6229}, 2013.

\bibitem{SouHub14}
Daniel Soudry, Itay Hubara, and Ron Meir.
\newblock Expectation backpropagation: Parameter-free training of multilayer
  neural networks with continuous or discrete weights.
\newblock In Zoubin Ghahramani, Max Welling, Corinna Cortes, Neil~D. Lawrence,
  and Kilian~Q. Weinberger, editors, {\em NIPS}, pages 963--971, 2014.

\bibitem{InceptionV4}
Christian Szegedy, Sergey Ioffe, and Vincent Vanhoucke.
\newblock Inception-v4, inception-resnet and the impact of residual connections
  on learning.
\newblock {\em CoRR}, abs/1602.07261, 2016.

\bibitem{mnasnet}
Mingxing Tan, Bo Chen, Ruoming Pang, Vijay Vasudevan, and Quoc~V. Le.
\newblock Mnasnet: Platform-aware neural architecture search for mobile.
\newblock {\em CoRR}, abs/1807.11626, 2018.

\bibitem{spse1992curves}
SPSE the Society~for Imaging~Science, Technology, Society of~Photo-optical
  Instrumentation~Engineers, and Technical~Association of~the Graphic~Arts.
\newblock {\em Curves and Surfaces in Computer Vision and Graphics}.
\newblock Number v. 1610 in Proceedings of SPIE--the International Society for
  Optical Engineering. SPIE, 1992.

\bibitem{FBnet}
Bichen Wu, Xiaoliang Dai, Peizhao Zhang, Yanghan Wang, Fei Sun, Yiming Wu,
  Yuandong Tian, Peter Vajda, Yangqing Jia, and Kurt Keutzer.
\newblock Fbnet: Hardware-aware efficient convnet design via differentiable
  neural architecture search.
\newblock {\em CoRR}, abs/1812.03443, 2018.

\bibitem{Shift}
Bichen Wu, Alvin Wan, Xiangyu Yue, Peter~H. Jin, Sicheng Zhao, Noah Golmant,
  Amir Gholaminejad, Joseph Gonzalez, and Kurt Keutzer.
\newblock Shift: {A} zero flop, zero parameter alternative to spatial
  convolutions.
\newblock {\em CoRR}, abs/1711.08141, 2017.

\bibitem{QCNN}
Jiaxiang Wu, Cong Leng, Yuhang Wang, Qinghao Hu, and Jian Cheng.
\newblock Quantized convolutional neural networks for mobile devices.
\newblock {\em CoRR}, abs/1512.06473, 2015.

\bibitem{NetAdapt}
Tien{-}Ju Yang, Andrew~G. Howard, Bo Chen, Xiao Zhang, Alec Go, Mark Sandler,
  Vivienne Sze, and Hartwig Adam.
\newblock Netadapt: Platform-aware neural network adaptation for mobile
  applications.
\newblock In {\em ECCV}, 2018.

\bibitem{ShuffleNet2017}
Xiangyu Zhang, Xinyu Zhou, Mengxiao Lin, and Jian Sun.
\newblock Shufflenet: An extremely efficient convolutional neural network for
  mobile devices.
\newblock {\em CoRR}, abs/1707.01083, 2017.

\bibitem{zhao2017pyramid}
Hengshuang Zhao, Jianping Shi, Xiaojuan Qi, Xiaogang Wang, and Jiaya Jia.
\newblock Pyramid scene parsing network.
\newblock In {\em CVPR}, 2017.

\bibitem{Increment_quant}
Aojun Zhou, Anbang Yao, Yiwen Guo, Lin Xu, and Yurong Chen.
\newblock Incremental network quantization: Towards lossless cnns with
  low-precision weights.
\newblock {\em CoRR}, abs/1702.03044, 2017.

\bibitem{DoReFa-Net}
Shuchang Zhou, Zekun Ni, Xinyu Zhou, He Wen, Yuxin Wu, and Yuheng Zou.
\newblock Dorefa-net: Training low bitwidth convolutional neural networks with
  low bitwidth gradients.
\newblock {\em CoRR}, abs/1606.06160, 2016.

\bibitem{NAS_reinforcement}
Barret Zoph and Quoc~V. Le.
\newblock Neural architecture search with reinforcement learning.
\newblock {\em CoRR}, abs/1611.01578, 2016.

\bibitem{LearningToLearnScale}
Barret Zoph, Vijay Vasudevan, Jonathon Shlens, and Quoc~V. Le.
\newblock Learning transferable architectures for scalable image recognition.
\newblock {\em CoRR}, abs/1707.07012, 2017.

\end{thebibliography}
}

\appendix
\section{Performance table for different resolutions and multipliers}
We give detailed table containing multiply-adds,  accuracy, parameter count and
latency in Table \ref{tab:v3_stats}.
\begin{table}
    \centering
    \begin{tabular}{l|c|c|c|c}
\toprule[0.2em]
Network & Accuracy & Madds & Params & P-1 \\
&          & (M)   & (M) & (ms) \\
\hline
large 224/1.25  & 76.6 & 356 &  7.5 & 77.0 \\
large 224/1.0  & 75.2 & 217 &  5.4 & 51.2 \\
large 224/0.75  & 73.3 & 155 &  4.0 & 39.8 \\
large 224/0.5  & 68.8 & 69 &  2.6 & 21.7 \\
large 224/0.35  & 64.2 & 40 &  2.2 & 15.1 \\
large 256/1.0  & 76.0 & 282 &  5.4 & 65.6 \\
large 192/1.0  & 73.7 & 160 &  5.4 & 38.0 \\
large 160/1.0  & 71.7 & 112 &  5.4 & 27.8 \\
large 128/1.0  & 68.4 & 73 &  5.4 & 17.8 \\
large 96/1.0  & 63.3 & 43 &  5.4 & 12.5 \\
small 224/1.25  & 70.4 & 91 &  3.6 & 23.6 \\
small 224/1.0  & 67.5 & 57 &  2.5 & 15.8 \\
small 224/0.75  & 65.4 & 44 &  2.0 & 12.8 \\
small 224/0.5  & 58.0 & 21 &  1.6 & 7.7 \\
small 224/0.35  & 49.8 & 12 &  1.4 & 5.7 \\
small 256/1.0  & 68.5 & 74 &  2.5 & 20.0 \\
small 160/1.0  & 62.8 & 30 &  2.5 & 8.6 \\
small 128/1.0  & 57.3 & 20 &  2.5 & 5.8 \\
small 96/1.0  & 51.7 & 12 &  2.5 & 4.4 \\
\bottomrule
\end{tabular}
\caption{Floating point performance for Large and Small V3 models. P-1  corresponds to large single core performance on Pixel 1. }
\label{tab:v3_stats}
\end{table}
\end{document}